# Carbon Emission Prediction on the World Bank Dataset for Canada

Pandit Deendayal Energy University

Aman Desai Shyamal Gandhi Sachin Gupta Manan Shah Samir Patel

## ABSTRACT

The continuous rise in CO2 emission into the environment is one of the most crucial issues facing the whole world. Many countries are making crucial decisions to control their carbon footprints to escape some of their catastrophic outcomes. There has been a lot of research going on to project the amount of carbon emissions in the future, which can help us to develop innovative techniques to deal with it in advance. Machine learning is one of the most advanced and efficient techniques for predicting the amount of carbon emissions from current data. This paper provides the methods for predicting carbon emissions (CO2 emissions) for the next few years. The predictions are based on data from the past 50 years. [7] The dataset, which is used for making the prediction, is collected from World Bank datasets. This dataset contains CO2 emissions (metric tons per capita) of all the countries from 1960 to 2018. Our method consists of using machine learning techniques to take the idea of what carbon emission measures will look like in the next ten years and project them onto the dataset taken from the World Bank's data repository.

The purpose of this research is to compare how different machine learning models (Decision Tree, Linear Regression, Random Forest, and Support Vector Machine) perform on a similar dataset and measure the difference between their predictions.

## 1. INTRODUCTION

With the increased use of carbon-containing fuels like coal, natural gas, and oil, the amount of carbon dioxide (CO2) emitted into the environment is also increasing. Carbon dioxide is a major contributor to the greenhouse effect and causes global warming, which eventually results in higher global temperatures, unpredictable weather patterns, and rising sea levels. The more carbon dioxide is emitted into the environment, the more heat will be trapped by these greenhouse gases, which will cause interruptions in our food chains and water supply patterns. Let's analyze the environmental impact of carbon emissions. They can also hinder global economic growth [11] [12]. Disrupted climate patterns will lead to a decrease in farming efficiency, which will damage the global production and supply chain architecture will then result in a rise in inflation and other economic crises. In the light of the catastrophic events caused by increased carbon emissions, many countries across the world are already taking initiatives to control their carbon footprints. In order to set appropriate policies to control the carbon footprint, carbon emission prediction has become a very meaningful approach. It can help to get an intuition about the future carbon-emission patterns, which can help countries to take relevant steps today to control the emissions in the future. Based on the prediction patterns, different industries can use innovative approaches to utilize their resources properly and decrease carbon emissions.

There are several methods to forecast carbon emissions. *[8] includes several traditional and advanced computer-based techniques, such as Artificial Neural Networks (ANN), Pinch Analysis, and Generated Coal Analyses (GCA), that have been used to predict the amount of carbon emission for the past decade.* In the past few years, there has been a rise in the various types of machine learning approaches in almost every field. Machine learning is one of the best techniques for forecasting the amount of carbon emissions. There has been a lot of research done to predict carbon emissions with the help of various machine learning techniques [9] and [10]. We can train the model to learn different patterns from the carbon emission data currently available, based on which it will give us an estimate of the amount of carbon emissions in the upcoming years.

The proposed method of this project compares different machine learning models. Our research consists of predicting carbon emissions for the upcoming years by using the provided dataset. The carbon emission dataset Carbon dioxide emissions per capita in metric tons for each country Up to 1990, the data was gathered from the Carbon Dioxide Information Analysis Center, and after 1990, it is CAIT data: Climate Watch 2020: GHG Emissions. World Resources Institute, Washington, DC. [7] For the purpose of our research, we have taken the carbon emissions of Canada. In 2018, Canada ranked as the 10th GHG emitting country or region. Canada is one of the countries that participated in the Paris Agreement [13] with 194 other countries to reduce the amount of carbon emitted and fight climate change. Since then, Canada has committed to reducing carbon emissions. This paper compares machine learning models like linear regression, decision trees, etc. Each of these models predicts carbon emissions in the form of $CO_2$ emissions from 2019 to 2030 for Canada. The results of these models are stored in one table to compare the results. In our dataset, we have taken the data from Canada to perform the prediction.

## 2. **RELATED WORKS**

In the research, Ümit Ağbulut[1] aims to analyze the trends in $CO_2$ emissions in Turkey and predict the upcoming energy demand. The paper used a combined dataset of the data from the World Bank (GDP, Population, CO2) and the Turkish Statistical Institute models (vehicle kilometer data). Machine learning models, for example, SVM, ANN, and Deep Learning models, are among the few models that are utilized and compared using six different statistical metrics. As shown in the paper, the ANN and machine learning models perform very similarly to actual results compared to the DL model. The combined prediction models show that $CO_2$ emissions in Turkey through the transportation sector are rising at an annual growth rate between 0.7% and 3.65%. A machine learning technique was advanced by Leerbeck et al.[2] to predict the intensity of $CO_2$ emissions from European power networks in relation to the need for flexible power supplies. The study used a dataset of almost 473 input variables, including electricity generation, demand, imports, weather, and so on. Using LASSO and an enhanced feature selection approach, the number of variables has been decreased to fewer than 30. Three-way backward models are combined with Softmax to get the final model. Depending on the weather prediction, predictability errors in the output range from 0.095 to 0.183, and in distant places, 0.029–0.160. Without comprehensive information about the location or other properties, the improved approach may be employed in any European electric bidding area network.

In their study Hamrani et al.[3] investigates emissions of greenhouse gasses from agricultural locations in Quebec, Canada. To forecast future GHG emissions, many types of machine learning models were applied, including conventional regression, shallow learning, and deep learning. The information was gathered over five years throughout Quebec's agricultural districts. It looked at $CO_2$ and nitrous oxide emissions from the soil. The predictive study, which included statistical comparisons and negative confirmation of $CO_2$ and N2O fluctuations, revealed that the LSTM model outperformed ML models with high R coefficients. This research suggests that LSTM might be a valuable and important method for predicting future GHG emissions. According to Xiaodong Li et al.[4], $CO_2$ emissions have increased dramatically as a result of the recent boom in transportation. Using a variety of datasets, the research examines the changes in $CO_2$ emissions in the top 30 $CO_2$ generating nations. World Development Indicators, the International Organization of Motor Vehicle Manufacturers, and the United Nations

Department of Economic and Social Affairs all contributed data to the final dataset. Features such as socioeconomic, transportation, and $CO_2$ emission metrics are included in the dataset. To forecast future $CO_2$ emissions, this study employs conventional least squares, SVM, and gradient boosting regression models. During the assessment, a few statistical measures, such as MAE, rRMSE, MAPE, R2, and N-fold cross-validation, are employed to evaluate these models. In the research done by Maksymilian Mądziel et al.[5], they created a macroscale $CO_2$ emission model for fully hybrid vehicles. It helps in estimating carbon emissions from these vehicles using collected parameters like velocity, acceleration, etc. The data is collected with the correct use of the PEMS system. Comparisons between different machine learning models like linear regression, gaussian process regression, cubic support vector machine, artificial neural network, etc. were performed using statistical metrics like RMSE, MAE, MSE, and R2. Among these machine learning models, the Gaussian Process Regression model best fits the $CO_2$ emission model creation method. This comparison can be used to create new microscale models for hybrid vehicles.

Zhao W et al.[14] built a technique to assess the co2 emitted by the power generation industry in China more precisely. The proposed methodology uses a dataset containing factors like population of the country, GDP/capita, consumption of coal and proportion of Thermal Power. The STIRPAT model is being used to analyze the factors affecting the $CO_2$ emission. After this, a combination of Ridge Regression and Welvet Neural Network and Gaussian Cuckoo search algorithm forecasts the factors affecting the $CO_2$ emission. $CO_2$ emission is predicted by plugging those predicted factors into the RR-STIRPAT model. Obtained results of prediction of $CO_2$ emission are comparatively more precise than provided by other techniques. Although the results obtained by this method is good, the limited availability of data restricts the application of advanced algorithms. Mardani et al.[15] presents an ensemble adaptive neuro fuzzy inference system (ANFIS) in order to study the neuro fuzzy model used to predict and analyze the interactions among economic growth, renewable energy usage, and $CO_2$ emissions in G8 + 5 countries. This predictive strategy generates ambiguous rules in the real-world Global Development Index database and typically correlates between input and output parameters. All nations except Brazil have a distorted U-shape according to the Sasabuchi – Lind – Mehlum WDI test. The proposed soft-emitting computer system to measure $CO_2$ emissions proved to be effective. To find out the correlation findings of the ANFIS study combination with the Environmental Kuznets Curve (EKC) hypothesis, the Sasabuchi – Lind – Mehlum (SLM) U. study was used.

In their study, Lazim Abdullah et al.[8] examined a literature on $CO_2$ emissions forecasting and estimating methodologies. They gathered and examined related papers published in international journals between 2003 and 2013 to find answers to these two questions. I What were the most prevalent approaches? (ii) What factors were investigated on a regular basis? Based on the general observations from the scholarly works, certain adjustments and prospective future work are recommended. According to the findings, energy consumption is the most common source of $CO_2$ emissions, followed by fuel burning, notably with fossil fuels, and so on. This study supports the idea that socioeconomic issues are the primary drivers of $CO_2$ emissions. This research not only shows which artificial intelligence approaches are the most popular, but it also shows how academics and policymakers may use the methods effectively. In another study done by Pao et al.[16], links among pollutant emissions, usage of energy, and production in Brazil during the period 1980–2007 were investigated. During the years 2008–2013, the Gray (GM) model was used for three variable predictions. It seems that carbon emission results from energy and inelastic production over time. This may be the result of uncontrolled land use and deforestation in Brazil, which is a measure of the country's emissions. Brazil must pursue a two-pronged approach to increasing investment in energy infrastructure and increasing energy efficiency to reduce pollution and prevent negative impact on economic growth. With less than 3% MAPEs, all the best GMs and ARIMA demonstrate outstanding speculative performance.

A trend analysis method was utilized to measure and anticipate energy-related $CO_2$ emissions in the study[17].To begin, the top 25 countries' $CO_2$ emissions, as well as worldwide total $CO_2$ emissions, are examined from 1971 to 2007. Because they lacked statistical significance, regression experiments with R2 values of less than 0.94 were eliminated from the procedure. There are statistically significant trends in India, South Korea, Iran, Mexico, Australia and the whole world. According to the conclusions of the study, the models for those nations might be used to anticipate future $CO_2$ emissions. The agreement between computed and expected $CO_2$ emissions is acceptable. Based on the Iskandar Development Region (IDR) Malaysian study, Kean et al.[18] used the FML Model to mimic city equivalent $CO_2$ emissions data to improve data access, integration, and analysis of gas emissions in developing countries at the site related to $CO_2$ emissions can be detected or immature. The basis for the established FML Model was laid out in the first phase of this study. Implemented simulations of energy consumption and $CO_2$ emitted and given to IDR by 2050. If current trends in socio-economic growth continue, IDR emissions could reach 44 million tons by 2050, which is seven times more than the 2005 level, according to the study. It also discusses the capabilities of the FML model, including its ability to replicate energy use at city level and CO2 emission patterns, and its potential to act as a decision-making tool in an urban planning process by providing capacity assessments to other urban policies.

## 3. PROPOSED METHODOLOGY

The project's basic idea was to understand carbon emission trends over decades of data and develop a predictive model that can efficiently and accurately predict emission numbers for the coming decades. The first step for this analysis was to gather the data and understand the parameters that are foundational to the predictive model. Quantitative data for carbon emission trends was collected from the World Bank's global public repository. [7]

The obtained dataset consisted of alphabetically sorted 60 years of carbon emissions numbers for all of the 266 countries (i.e., 266 rows x 60 columns). [7] The dataset required a few data preparation and preprocessing steps. Data for a particular country was extracted from the dataset. The numerical data present in the dataset was then visualized using the basic python library of MaPlotLib to understand the trends in more detail. Based on that, the further course of research and the algorithms were decided for the data.

For this research purpose, four machine learning algorithms are used to create four different predictive models, which are Decision Tree, Random Forest, Linear Regression, and Support Vector Machine. A comparative study amongst all the four machine learning algorithms for their application in predictive analysis is also carried out as part of the research project. The use of four algorithms exhibited noteworthy outcomes. It was observed that the correlations between derived results and expected CO2 emissions from various scenarios were satisfactory.

## 4. IMPLEMENTATION DETAILS

The execution of the proposed method is mainly done in three different parts.

### 4.1. Data Preparation:

The CO2 emission dataset contains multiple columns which include carbon emissions for every country from 1960 to 2018. For this research, we have used Canada's carbon emission data for all the given years. Therefore, we have extracted one country's data from a combined dataset using the pandas dataframe object in Python. The plot [Fig 1] for carbon emissions from 1960 to 2018 in Canada.

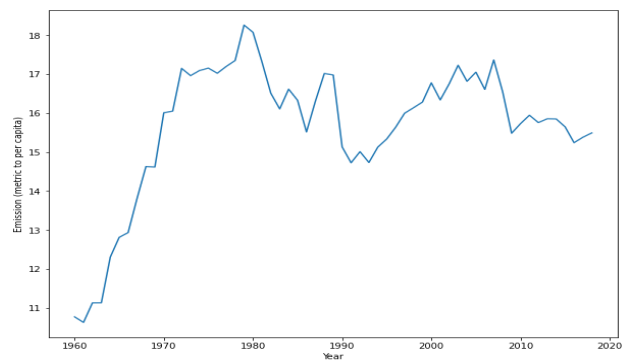

**Figure 1: Carbon Emissions in Canada (metric tons per capita) from 1960 to 2018.**

## 4.2. Model Training and Testing

In order to predict carbon emissions for the next few years, we have trained different machine learning models:

- **Linear Regression**

    It is mainly a statistical method which defines the relation between dependent and independent variables. In machine learning we use Linear Regression algorithm to train the model to learn this relation between the variables using something called as loss function. To simply put we will try to minimize the euclidean distance between predicted values and actual values.

    Hypothesis function for Linear Regression:[23]

    $$y = \theta_1 + \theta_2.x$$

    Here y is a dependent variable and x is a dependent variable. The linear regression algorithm will predict the value of y having minimum euclidean distance between the prediction and the real value. As a result, it is critical to update the 1(intercept) and 2(coefficient of x) values in order to find the ideal value that minimizes the square error values between the predicted and real y values, and this is carried out by using a loss function(J):

    $$J = \frac{1}{n} \sum_{i=1}^{n} (pred_i - y_i)^2$$

- **Random Forest Regression**

    It is a ML technique which uses multiple decision trees to reach out to the final output. In addition to that this technique of machine learning also uses Bootstrap and Aggregation which is sometimes known as bagging. Both regression and classification problems can be solved using the Random Forest algorithm. The objective of using this algorithm is that it provides the benefits of Decision Trees and minimizes the inaccuracy caused by using individual decision trees by aggregating through numerous decision trees and combining all the results into the final outcome.

- **Decision Tree Regression**

    Decision tree regression examines the properties of an object and trains a tree-based model to predict future data and create meaningful continuous output. The output/result is not discrete in the sense that it is not represented solely by a discrete, known set of numbers or values.[19]

    The link between the outcome y and the attributes x is described by the formula below.[23]

    $$\hat{y} = \hat{f}(x) = \sum_{m=1}^{M} c_m I\{x \in R_m\}$$

    Here,
    $R_m$ = leaf nodes,
    $I_x R_m$ = identity function

    The output of this function will be 1 if x is a leaf node otherwise the output will be 0. Now, if the Rl x is a leaf node, it will result in y = cl. Here cl represents the average of all the training instances in Rl leaf node.

- **Support Vector Machine**

    Both classification and regression problems can be solved with the help of

SVM algorithm. It falls under the category of supervised machine learning techniques. The algorithm finds an N-dimensional space hyperplane which classifies data points. The number of features determines the size of the hyperplane. If there are 2 input features then the hyperplane will be a line, similarly if there are 3 input features then the hyperplane will be a 2-dimensional plane. When there are more than three characteristics, it gets harder to imagine.

Let's say we have a set of training data with $x_n$ which is a multivariate collection of N observations and $y_n$ being the collected response values.

$$f(x) = x'\beta + b,$$

Find f (x) with the smallest norm value and ensure that it is as flat as feasible. This is framed as a convex optimization problem with the goal of minimizing the total cost.[24]

$$J(\beta) = \frac{1}{2}\beta'\beta$$

subject to the value of all residuals being smaller than; or, in equation form:

$$\forall n : |y_n - (x_n'\beta + b)| \leq \varepsilon.$$

It can happen that there doesn't exist any function f(x) which can satisfy all these restrictions everywhere. To deal with otherwise infeasible restrictions, variables n and *n which are also known as slack variables are introduced for each point. This technique is analogous to the "soft margin" notion in SVM classification. As the slack variables enable regression errors up to the value of ξn and ξ*n, [22]

$$J(\beta) = \frac{1}{2}\beta'\beta + C\sum_{n=1}^{N}\left(\xi_n + \xi_n^*\right)$$

subject to:

$$\forall n : y_n - (x_n'\beta + b) \leq \varepsilon + \xi_n$$

$$\forall n : (x_n'\beta + b) - y_n \leq \varepsilon + \xi_n^*$$

$$\forall n : \xi_n^* \geq 0$$

$$\forall n : \xi_n \geq 0.$$

For training and testing purposes, the dataset is split into a 90-10 ratio. 90% of all available datasets are used to train the models, with the remaining 10% being used to test the accuracy of trained models. The available Python library for machine learning models: sklearn, is used for the models. The accuracy is measured for all four models to compare the outcomes.

**4.3 Emission Prediction**

In order to predict the emissions for the future, we have used the trained machine learning models from the previous step. The results are generated using the sklearn libraries' pre-defined predict method. The outcomes are collected and stored in a combined table to compare the results generated by each model. Line charts have been plotted in order to visualize the predictive results of each of the four machine learning algorithms.

**5. RESULT ANALYSIS**

In this research, we used four different machine learning algorithms to train the model. By using these machine learning techniques, we have predicted the value of CO2 emissions for the next 10 years.

We have evaluated the test dataset on different machine learning models and then compared all the algorithms.[Fig 4] Among all the machine learning algorithms we have used, the Random Forest has given the best performance and accuracy of 83% on the test dataset.

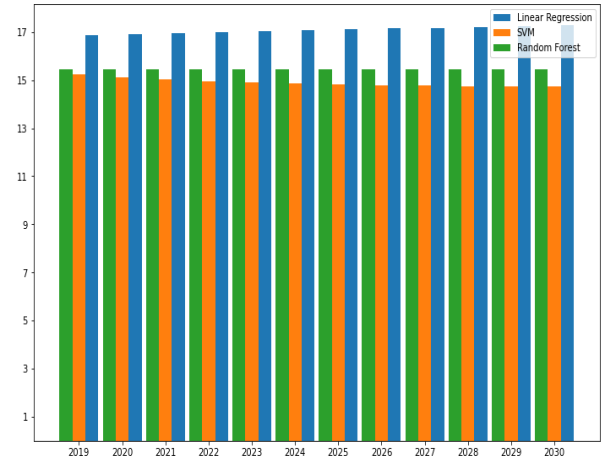

**Fig 2**: Prediction Using Linear Regression, SVM and Random Forest

After training the dataset, we have predicted the value of $CO_2$ emissions for the upcoming 10 years. Here is the table [Fig 3] of predicted values using different machine learning algorithms:

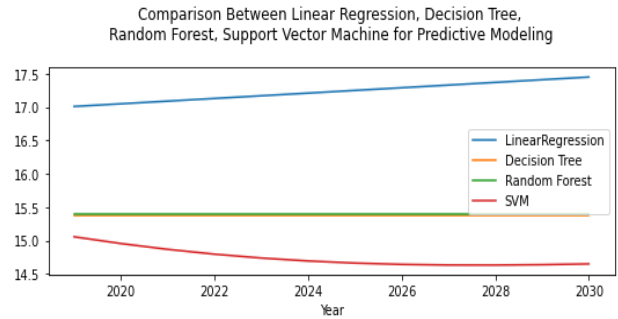

**Fig. 4**: Visualization of predicted values by four commonly used machine learning algorithms

## 6. **CONCLUSION**

As our research has demonstrated that the amount of $CO_2$ emissions in Canada was increasing for the first few decades, after that there has been a sudden decrease in the value. As our model has predicted, in the coming 10 years, there isn't going to be a significant variation in the values of $CO_2$ emissions. There are still other factors that contribute to carbon emissions and can affect the trends. Still, the outcomes from this study can give us the basic sense of future carbon emissions.

| Year | LinearRegression | Decision Tree | Random Forest | SVM |
|------|------------------|---------------|---------------|-----------|
| 2019 | 17.210947 | 15.385291 | 15.518 | 15.073555 |
| 2020 | 17.263663 | 15.385291 | 15.518 | 14.963301 |
| 2021 | 17.316378 | 15.385291 | 15.518 | 14.865553 |
| 2022 | 17.369093 | 15.385291 | 15.518 | 14.780883 |
| 2023 | 17.421808 | 15.385291 | 15.518 | 14.709383 |
| 2024 | 17.474523 | 15.385291 | 15.518 | 14.650720 |
| 2025 | 17.527238 | 15.385291 | 15.518 | 14.604215 |
| 2026 | 17.579953 | 15.385291 | 15.518 | 14.568916 |
| 2027 | 17.632669 | 15.385291 | 15.518 | 14.543678 |
| 2028 | 17.685384 | 15.385291 | 15.518 | 14.527239 |
| 2029 | 17.738099 | 15.385291 | 15.518 | 14.518282 |
| 2030 | 17.790814 | 15.385291 | 15.518 | 14.515501 |

**Fig. 3**: Carbon Emission Predicted values in Canada for the next decade (2019-2030)

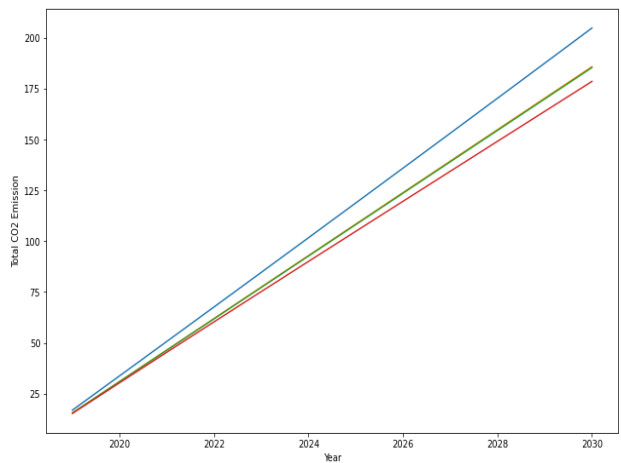

**Figure 5: ML models predict a consistent increase in CO2 emissions.**